\documentclass[11pt]{article}

\usepackage[preprint]{acl}

\usepackage{enumitem}
\usepackage{times}
\usepackage{latexsym}
\usepackage{amsmath}
\usepackage{multirow}
\usepackage{booktabs}
\usepackage{graphicx}
\usepackage{caption}
\usepackage{wrapfig}
\usepackage{hyperref}
\usepackage{url}
\usepackage{subcaption}
\usepackage[ruled,vlined]{algorithm2e}
\usepackage{amssymb}
\usepackage[T1]{fontenc}

\usepackage[utf8]{inputenc}

\usepackage{microtype}

\usepackage{inconsolata}

\usepackage{graphicx}

\title{Beyond Token-Level Policy Gradients for Complex Reasoning with Large Language Models}

\author{Mufan Xu$^{1\space}$, Kehai Chen$^{1\space\dagger}$, \textbf{Xuefeng Bai$^{1}$}, \textbf{Zhengyu Niu$^{2}$}, \textbf{Muyun Yang$^{1}$},\\ \textbf{Tiejun Zhao$^{1}$, Min Zhang$^{1}$}\\
  $^{1}$School of Computer Science and Technology, Harbin Institute of Technology, China \\$^{2}$Baidu Inc., Beijing, China\\
  \texttt{xmuffins0610@gmail.com}, \texttt{niuzhengyu@baidu.com}\\  \texttt{\{chenkehai,baixuefeng,yangmuyun,tjzhao,zhangmin2021\}@hit.edu.cn}}

\begin{document}
\maketitle
\def\thefootnote{$\dagger$}\footnotetext{Corresponding author.}
\renewcommand{\thefootnote}{\arabic{footnote}}
\begin{abstract}
Existing policy-gradient methods for auto-regressive language models typically select subsequent tokens one at a time as actions in the policy.
While effective for many generation tasks, such an approach may not fully capture the structure of complex reasoning tasks, where a single semantic decision is often realized across multiple tokens—for example, when defining variables or composing equations. This introduces a potential mismatch between token-level optimization and the inherently block-level nature of reasoning in these settings.
To bridge this gap, we propose \textbf{M}ulti-token \textbf{P}olicy Gradient \textbf{O}ptimization (\textbf{MPO}), a framework that treats sequences of $K$ consecutive tokens as unified semantic actions. 
This block-level perspective enables our method to capture the compositional structure of reasoning trajectories and supports optimization over coherent, higher-level objectives.
Experiments on mathematical reasoning and coding benchmarks show that MPO outperforms standard token-level policy gradient baselines, highlight the limitations of token-level policy gradients for complex reasoning, motivating future research to look beyond token-level granularity for reasoning-intensive language tasks. \footnote {All codes will be released upon acceptance.}
\end{abstract}

\section{Introduction}

Large language models (LLMs) have become the foundation of modern natural language understanding and generation~\citep{kumar2024large,jiang2024survey}, achieving remarkable results through large-scale pretraining and autoregressive modeling~\citep{tang2025thinking}. Recently, there has been increasing interest in leveraging policy gradient methods to further fine-tune these models, with the aim of enhancing their capacity for complex reasoning and long-horizon dependency modeling~\citep{ouyang2022training}. Among these methods, Proximal Policy Optimization (PPO) has emerged as a dominant framework, offering efficient and stable updates for reinforcement learning-based fine-tuning in LLMs~\citep{schulman2017proximal}. Other advanced methods, such as Group Relative Policy Optimization (GRPO) \citep{deepseekmath} and Decoupled Clip and Dynamic Sampling Policy Optimization (DAPO) \citep{yu2025dapo}, employ group-based and adaptive sampling strategies to address the challenges of structured reasoning and mathematical problem-solving.

\begin{figure*}
    \centering
    \includegraphics[width=1\linewidth]{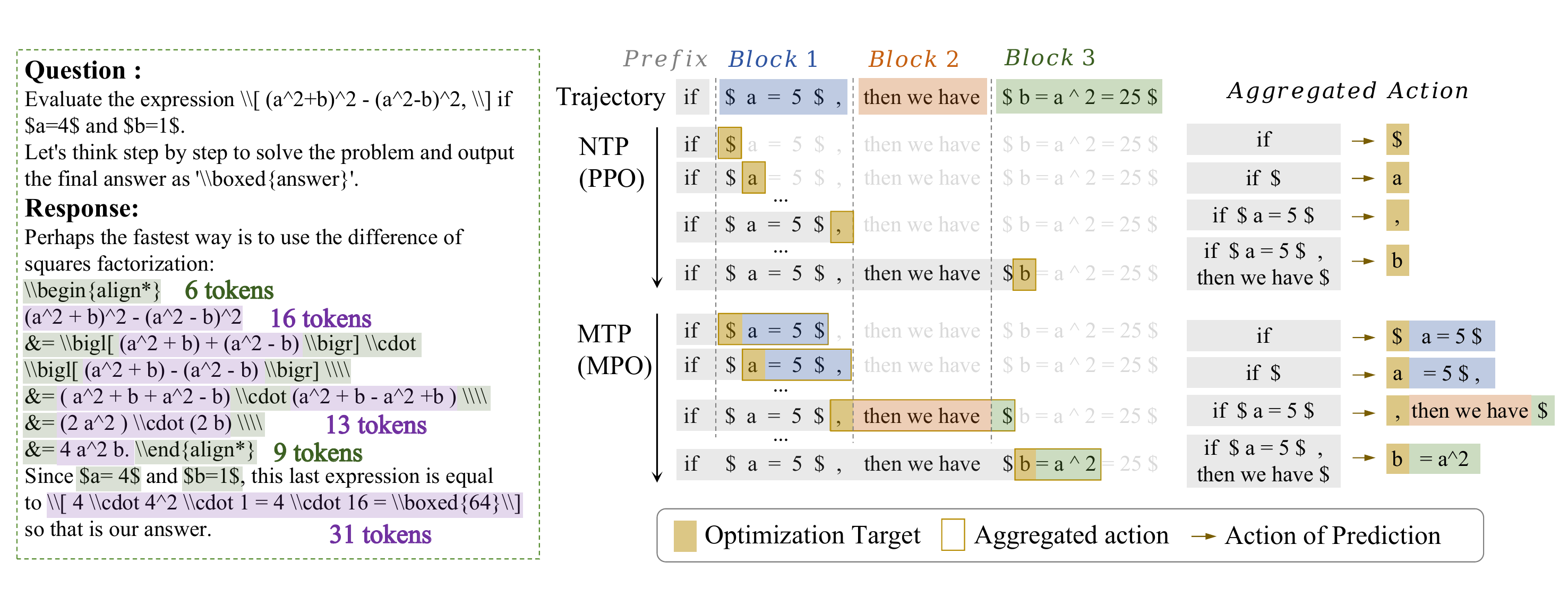}
    \caption{\textbf{Left}: In reasoning tasks such as mathematical problem-solving or code generation, the model’s decision process often spans across blocks of tokens—such as equations or functions—rather than being determined by each token independently; \textbf{Right}: illustration of token-level (NTP/PPO) vs. block-level (MTP/MPO) optimization. MPO aggregates K tokens as a semantically meaningful block for prediction and optimization, thereby better capturing sequence structure and long-range dependencies.}
    \label{fig:motivation}
\end{figure*}

However, as shown in Figure~\ref{fig:motivation} left, in complex reasoning tasks such as mathematical reasoning, decision-making (like defining variables or completing equations) evolves over semantic units which contains multiple tokens. Existing policy gradient techniques decompose these structured reasoning steps into a series of local token predictions, disrupting the coherence of the underlying semantic actions (Figure~\ref{fig:motivation} right).
This fundamental granularity mismatch calls for optimization frameworks that transcend the limitations of token-level actions, enabling more semantically meaningful and globally coherent reasoning abilities in large language models~\citep{mirzadehgsm,zhu2024benchmarking,zhang2025tokenization}. Overcoming this issue calls for training strategies that capture reasoning actions at a coarser, semantically meaningful level beyond single-token updates.

In this paper, we introduce \emph{Multi-token Policy Gradient Optimization} (MPO), a framework that incorporates block-level action into the policy gradient process. 
As illustrated in Figure~\ref{fig:motivation} right, instead of treating each token generation as an isolated action, MPO aggregates contiguous blocks of $K$ tokens and unites importance sampling ratios over them. MPO allows the model to consider multiple correlated tokens as a single semantic action, better preserving the internal structure of reasoning steps such as variable definitions, function calls, or equation formations. This block-level optimization encourages the policy to plan over meaningful reasoning segments rather than isolated symbols, thus maintaining consistency across intermediate decisions and improving global reasoning coherence. MPO only modifies the importance sampling ratio; it remains broadly compatible with existing policy-gradient frameworks and can be seamlessly integrated into contemporary LLM post-training pipelines. Our main contributions are:

\begin{enumerate}[itemsep=1pt, topsep=1pt, parsep=1pt, partopsep=1pt]
    \item We propose \textbf{Multi-token Policy Gradient Optimization (MPO)}, which enables the policy to optimize over structured reasoning units rather than isolated tokens.
    \item We incorporate structural multi-token optimization into the post-training stage of LLMs, revealing its potential to enhance reasoning coherence and informing new directions for policy-gradient research.
    \item MPO consistently outperforms token-level policy gradient baselines on mathematical reasoning and code generation benchmarks, demonstrating its effectiveness in improving reasoning ability during post-training.
\end{enumerate}

\section{Related Work}

\subsection{Multi‑Token Prediction with LLMs}
Multi-token prediction (MTP) is an extension of standard auto-regressive language modeling, where typically, the model is trained to predict only the immediate next token given the preceding context. Gloeckle et al.~\citep{gloeckle2024better} propose using multiple prediction heads to forecast several tokens from each context position, improving sample efficiency and inference speed on coding and generative benchmarks. This approach demonstrates stronger induction and reasoning ability gains, particularly at larger model scales like DeepSeek V3~\citep{liu2024deepseek}. Gerontopoulos et al.~\citep{gerontopoulos2025multi} extend this to freeze pre‑trained models and add learnable "register tokens" to support multi‑token prediction without modifying the backbone. 

While previous works on multi-token prediction enhanced LLM's sample efficiency and accelerated inference, our work extends the MTP concept from pre-training or inference use into the realm of policy-gradient optimization.

\subsection{Post‑Training RL Algorithms}
Reinforcement Learning from Human Feedback (RLHF) is an approach to align language models with human preferences by optimizing the policy using reinforcement learning on preference-labeled data. 
PPO~\citep{schulman2017proximal} introduced the clipped surrogate objective to stabilize updates, becoming the backbone of early RLHF such as InstructGPT~\citep{ouyang2022training}.  
Building on PPO, approaches such as VinePPO~\citep{kazemnejadvineppo} introduce intermediate rollouts to facilitate more accurate long-context credit assignment, thereby enhancing policy optimization in autoregressive frameworks.
Group Relative Policy Optimization (GRPO)~\citep{deepseekmath} computes advantages by comparing multiple sampled completions per prompt (group‑based advantage), eliminating the need for a value network and enhancing learning on reasoning tasks.  
DAPO (Decoupled Clip and Dynamic Sampling Policy Optimization)~\citep{yu2025dapo} further refines GRPO’s token‑level loss by allowing asymmetric clipping bounds (“clip‑higher”) and dynamic sampling to maintain a helpful gradient signal.  
VAPO~\citep{yue2025vapo} integrates value‑based methods into reasoning and RLHF, augmenting PPO variants with adaptive GAE and value pretraining for enhanced stability and performance.
GSPO~\citep{zheng2025group} (Group Sequence Policy Optimization) optimizes at the sequence level using importance sampling and clipping, improving stability and efficiency in policy optimization for LLMs. However, our MPO focuses on block-level actions, better aligning with the semantic reasoning units which are typically local blocks rather than entire sequences.
CISPO~\citep{chen2025minimax} (Minimax-style policy optimization) methods have been proposed mainly in adversarial language alignment, thus falling outside our comparison.

While prior work enhanced PPO variants via sampling tweaks, decoupled clipping, or value modeling, to our knowledge, none have incorporated \emph{multi-token action} in policy gradient optimization process as illustrated in Figure~\ref{fig:motivation}.

\section{Preliminaries}
\label{sec:preliminary}

We first review existing approaches to importance sampling in policy optimization—specifically PPO and its extensions such as GRPO—to clarify their formulation and highlight their limitations. This sets the stage for introducing our method, which extends importance sampling to account for longer-horizon behavior via multi-token predictions.

\subsection{Importance Sampling Strategy in Policy Optimization}

Proximal Policy Optimization (PPO)~\citep{schulman2017proximal} relies on importance sampling to support multiple epochs of minibatch updates with trajectories collected under a previous policy. The surrogate objective uses:
\begin{equation}
\begin{split}
r_t = \frac{\pi_{\theta}(a_t \mid s_t)}{\pi_{\theta_{\text{old}}}(a_t \mid s_t)},
\end{split}
\end{equation}
and the clipped PPO loss is
\begin{equation}
\begin{split}
J(\theta) = \mathbb{E}_t \left[ \min\big(r_t \hat{A_t}, \mathrm{clip}(r_t,\,1\pm\epsilon) \hat{A_t}\big) \right],
\end{split}
\end{equation}
where \(\hat{A_t}\) is the advantage estimate, and the clipping constrains large updates to maintain stability. Generalized Reward Policy Optimization (GRPO)~\citep{deepseekmath} modifies the importance sampling ratio by introducing group-relative advantage sampling. The same importance ratio \(r_t = \frac{\pi_\theta(a_t|s_t)}{\pi_{\theta_\text{old}}(a_t|s_t)}\) is used per sample, and the loss aggregates contributions averaged over sampled actions. Decoupled Clip and Dynamic Sampling Policy Optimization (DAPO) similarly builds on this framework by adapting the clipping threshold and employing rejection sampling, but does not alter the fundamental importance sampling ratio.

However, in the context of language modeling, where accurate semantics and the maintenance of long-range dependencies depend on structural token blocks, the commonly used token-level ratios fail to capture behavioral differences or dependencies that span multiple steps, and the standard per-step ratio may become highly volatile~\citep{metelli2020importance,papini2024policy}.

\begin{figure*}[ht]
    \centering
    \includegraphics[width=1\linewidth]{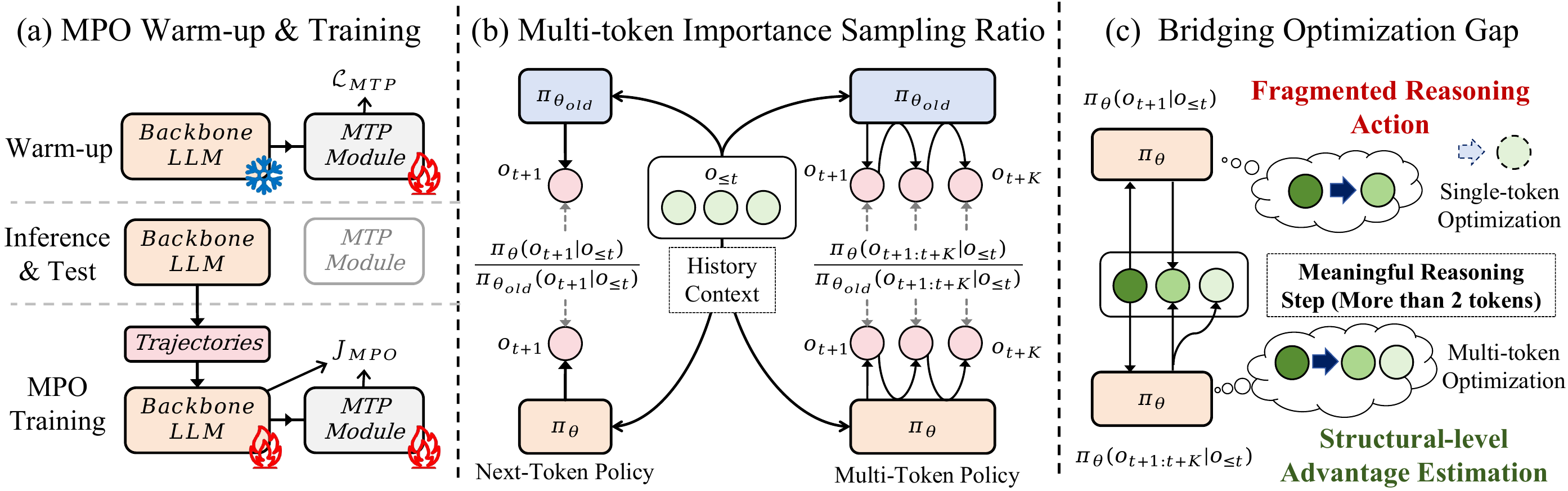}
    \caption{(a) Demonstration of the implementation of MPO warm-up and training process; (b) illustration of the united importance sampling ratio proposed in MPO method; (c) Comparison between single-token and multi-token optimization, multi-token optimization jointly models contiguous tokens as a structural reasoning action.}
    \label{fig:method}
\end{figure*}

\subsection{Multi‑Token Prediction Mechanism}
\label{sec:mtp_warmup}

To address the limited expressiveness of token-level representations, we turn to a multi-token representation mechanism. The MTP implementation introduced in DeepSeek‑V3/R1~\citep{liu2024deepseek} allows the model at each position \(t\) to predict up to \(K\) tokens \(o_{t+1}, o_{t+2}, \dots, o_{t+K}\) through a sequence of MTP modules that preserve causal consistency (we provide detailed implementation of MTP modules in Appendix~\ref{app:implementation}). Specifically, the probability distribution on position $t+k$ is:
\begin{equation}
\begin{split}
h_{t}^{k} = L_k\bigl(M_k\bigl(h_t^{k-1},\mathrm{Emb}(o_{t+k-1})\bigr)\bigr),
\end{split}
\end{equation}
where \(h_t^k\in\mathbb{R}^h\) is the hidden state of the $k^{th}$ MTP module, note that $h_t^0$ represents the hidden state output of the backbone model for token $o_t$. \(M_k\) is a learned projection layer, \(L_k\) is a Transformer decoder block, and Emb\((o_{t+k})\) is the embedding of token $o_{t+k}$. Each module outputs with an individual softmax prediction for the probability distribution of the $t+k+1$ token of the sequence:
\begin{equation}
\begin{split}
p(o_{t+k} \mid q,o_{1:t+k-1}) = \mathrm{LM\_Head}(h_t^k),
\end{split}
\end{equation}
for \(k = 2,\dots,K\). In supervised fine-tuning settings, the MTP objective is defined as:
\begin{equation}
\label{eq:mtp_loss}
\begin{split}
\mathcal{L}_{\text{MTP}} = -\sum_{k=2}^{K}\alpha_{k}\log p(o_{t+k} \mid q,o_{1:t+k-1}).
\end{split}
\end{equation}
The loss of the MTP module is assigned a decaying weight $\alpha_{k}$ to simulate the diminishing weights of tokens. Although multi-token prediction has been adopted in various pre-training and fine-tuning settings to improve generation efficiency and performance, its integration into reinforcement-based post-training remains limited. Existing policy‑gradient approaches, including PPO, GRPO, and DAPO, remain confined to token‑level updates and overlook the structured, multi‑step action patterns inherent to reasoning‑oriented generation.

Following previous works~\citep{cai2024medusa,ankner2024hydra}, MPO first initializes the MTP modules using the last layer of the backbone model, then warm-up these modules using the objective (\ref{eq:mtp_loss}) to ensure the quality of multi‑token prediction, see Appendix~\ref{app:implementation} for more details.

\section{Multi-token Objective}
\label{sec:future}

To address the limitations discussed in Section~\ref{sec:preliminary}, we move beyond the conventional token-level optimization paradigm and introduce a framework that directly incorporates block-level semantic structure into policy gradient updates, thereby better aligning model optimization with the demands of complex reasoning.

\subsection{From Single-Token to Multi-Token}

Let $q_i$ denote the prompt (input) for trajectory $i$, and $o_{i,1:t}$ denote the sequence of generated tokens up to position $t$ for trajectory $i$. The usual per-token importance sampling ratio $r_{i,t}$ for trajectory $i$ at position $t$ is defined as: 
\begin{equation}
\begin{split}
r_{i,t}(\theta) \;=\;\frac{\pi_\theta( o_{i,t+1}\mid q_i,o_{i,1:t} )}{\pi_{\theta_\text{old}}( o_{i,t+1}\mid q_i,o_{i,1:t} )}.
\end{split}
\end{equation}
With the help of multi-token prediction modules, we can replace the importance sampling ratio with an aggregated ratio over spans of length K (as shown in Figure~\ref{fig:method}(b)):  
\begin{equation}
\begin{split}
r_{i,t}(\theta) \;=\;\frac{\pi_\theta\bigl(o_{i,t+1:t+K}\mid q_i,o_{i,1:t}\bigr)}{\pi_{\theta_\text{old}}\bigl(o_{i,t+1:t+K}\mid q_i,o_{i,1:t}\bigr)}.
\end{split}
\end{equation} 
The new objective then sums (or clips) these block-level ratios times advantages. This encourages coherent multi-token patterns rather than independent next-token moves. 
For example, when constructing the deepseek-r1 reasoning model, the pre-training stage cost around 90\% of the total cost, and the post-training (supervised fine-tuning and then reinforcement learning alignment) cost the remaining 10\% budget. For those pretrained LLMs that did not use MTP loss during their pre-training process, we provide a solution to update the model's multi-token prediction ability in a relatively lower-cost post-training way.

\subsection{Multi-Token Policy-Gradient Objective}  
\label{sec:mul-obj}

Let each trajectory \(i\) produce a sequence of generated tokens using the original auto-regressive strategy 
\(o_i = \left(o_{i,1}, o_{i,2}, \dots, o_{i,|o_i|}\right)\).
After generating the base sequence $o_{i,1:t}$ for prompt $q_i$, we compute the multi-token importance sampling ratio by sampling $K$ additional tokens in an auto-regressive fashion. Specifically, at each position $t$, for each $n \in [1, K]$, the generation of token $o_{i,t+n}$ is conditioned on the entire prefix $o_{i,1:t+n-1}$, meaning each new token is predicted based on all previously generated tokens, including those just sampled within the span. The revised multi-token importance sampling ratio is thus formulated as:
\begin{equation}
\label{eq:mtp_ratio}
\begin{split}
R_{i,t}^{(K)}(\theta)
= \prod_{n=1}^{K}
\frac{\pi_\theta\bigl(o_{i,t+n} \mid o_{i,1:t+n-1}\bigr)}
{\pi_{\theta_{\text{old}}}\bigl(o_{i,t+n} \mid o_{i,1:t+n-1}\bigr)}.
\end{split}
\end{equation}
Though equation~\ref{eq:mtp_ratio} can effectively reduce bias by taking more future actions into consideration (see mathematical derivations in Appendix~\ref{app:math_prove}), such a production form of the importance ratio tends to dramatically increase the variance of sampling ratios, which further leads to a large clip fraction and hinders the optimization process in practice.
Inspired by the Log‑COP‑TD method~\citep{hallak2017consistent}, we adopt an alternative trade‑off formulation to control variance, replacing the product of ratios with a weighted log‑sum:
\begin{equation}
\begin{split}
\widetilde{R}_{i,t}^{(K)}(\theta)&= 
\exp\!\left(\sum_{n=1}^{K} \beta_n 
\log r_{i,t+n}(\theta)\right),\\ 
r_{i,t+n}(\theta) &= \frac{\pi_\theta\!\left(o_{i,t+n} \mid o_{i,1:t+n-1}\right)}
               {\pi_{\theta_{\text{old}}}\!\left(o_{i,t+n} \mid o_{i,1:t+n-1}\right)},
\end{split}
\end{equation}
where ${\beta_n}$ are non‑negative step‑wise weights satisfying $\sum_{n=1}^K \beta_n = 1$. Based on the weight of the first MTP module, denoted as $\beta_2$, we define 
\begin{equation}
\label{eq:decay}
\begin{split}
\beta_k = \beta_2 \times \lambda^{k-2}, \quad k \ge 2; 0 \le \lambda \le 1,
\end{split}
\end{equation}
where $\lambda$ is a hyperparameter controlling the rate of information decay. 
This formulation applies a decaying weight to the $(k-1)^{\text{th}}$ MTP module, thereby incorporating the influence of multi-token information with diminishing strength. This design prioritizes the impact of nearer token predictions while still incorporating longer-horizon contributions in a controllable manner. We then propose the policy‐gradient surrogate objective of MPO, analogous to PPO but using the \(K\)-step ratio at each position $t$:
\begin{equation}
\begin{split}
&J_{MPO}(\theta) 
= \mathbb{E}_{q_i\sim D,\; {\{o_i\}}_{i=1}^G\sim\pi_{\theta_{old}}(.|q_i)} \\
&\Biggl[\frac{1}{\sum_{i=1}^{G}|o_i|}{\sum_{i=1}^{G}}\sum_{t=1}^{|o_i|}
   \min\Bigl( \widetilde{R}_{i,t}^{(K)}(\theta)\, \hat A_{i,t}, \\&\mathrm{clip}\bigl(\widetilde{R}_{i,t}^{(K)}(\theta),1-\epsilon_{low},1+\epsilon_{high}\bigr)\, 
   \hat A_{i,t} \Bigr) \Biggr],
\end{split}
\end{equation}
where \(\hat A_{i,t}\) is the estimated advantage at position \(t\). By jointly observing \(K\) consecutive tokens in a single step, the estimator encourages optimization to account for a broader, structurally coherent view at the level of multi‑token reasoning chunks (as illustrated in Figure~\ref{fig:method}(c)). In this paper, we focus on the experiments with $G=1$ and we set $\epsilon_{low} = \epsilon_{high}$. We also conduct experiments with MTP-based value estimation in MPO, see Appendix~\ref{app:value}.

\section{Experimental results}

In this section, we provide a brief overview of our experimental setup, including the baselines, evaluation methods, and training hyperparameter configurations. And then, we analyze the task performance of MPO and the effects of introducing multi-token information and the training efficiency.

\subsection{Experimental Settings}

\textbf{Model and Datasets.} We implement MPO on three widely used instruction‑tuned backbone models: Llama3.2‑1B‑Instruct~\citep{dubey2024llama}, DeepSeek‑Distilled‑Qwen2.5‑1.5B, and DeepSeek‑Distilled‑Qwen2.5‑7B~\citep{guo2025deepseek}, to evaluate its effectiveness across different architectures and model scales.
We assess MPO on two mathematical reasoning benchmarks of varying difficulty, GSM8K~\citep{cobbe2021training} (college‑level) and MATH~\citep{hendrycks2021measuring} (competition‑level), and further evaluate its code generation capability on the HumanEval benchmark~\citep{chen2021evaluating} to test its cross‑domain generalization. For HumanEval, we use coding benchmark MBPP~\citep{austin2021program} as the training set. See Dataset statistics in Appendix~\ref{app:data_details}.

\begin{figure*}[ht]
    \centering    
    \includegraphics[width=1\linewidth]{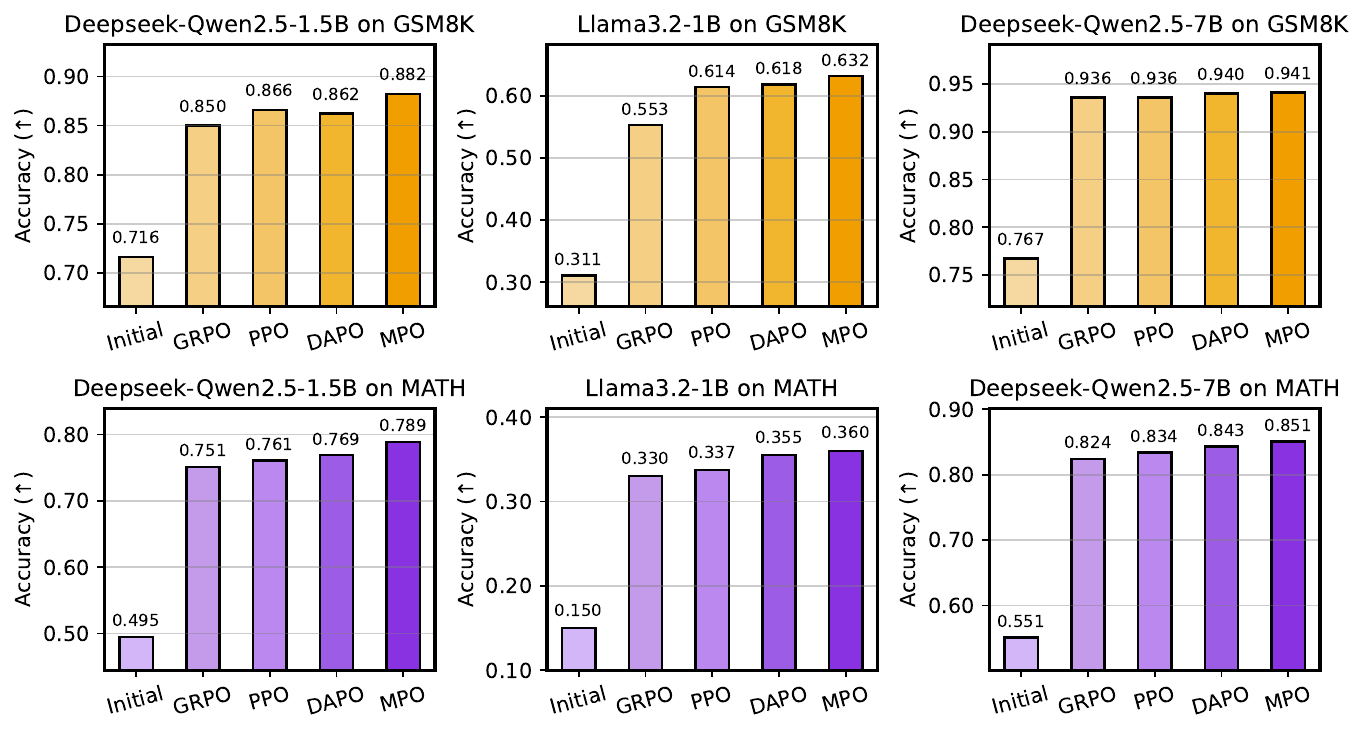}
    \caption{The performance of proposed MPO and baseline methods. MPO outperforms the baselines in most scenarios, demonstrating the effectiveness of aggregating block-wise information.}
    \label{fig:main_result}
\end{figure*}

\begin{table*}[ht]
    \centering
    \begin{tabular}{l|ccc|cc}
    \hline
    Model  &Zero-Shot & GRPO & DAPO & PPO & MPO (Ours) \\ \hline
    Llama3.2‑1B‑Instruct              & 0.354 & 0.372 & 0.396 & 0.390 & \textbf{0.403} \\ \hline
    DeepSeek‑Qwen2.5‑1.5B           & 0.451 & 0.591 & 0.603 & 0.598 & \textbf{0.640} \\ \hline
    DeepSeek‑Qwen2.5‑7B           & 0.689 & 0.811 & 0.817 & 0.805 & \textbf{0.841} \\ \hline
    \end{tabular}
    \caption{The performance of proposed MPO and baseline methods on coding task HumanEval, MPO consistently outperforms the baselines, demonstrating the effectiveness of aggregating block-wise actions during optimization.}
    \label{tab:humaneval}
\end{table*}

\noindent\textbf{Evaluation.} We use pass@1 metric to evaluate the models’ accuracy in answering mathematical questions and coding completions, which focuses on the correctness of the final answer provided by the model upon completion of the reasoning process. For math problems, to unify the evaluation process, we use the answering format of \verb|"\boxed{answer}"|. We select the most commonly used token-wise policy optimization strategies: PPO~\citep{schulman2017proximal}, GRPO~\citep{deepseekmath}, and DAPO~\citep{yu2025dapo}. In this paper, we mainly discuss the implementation and results of MPO based on PPO.






\noindent\textbf{Implementation.} As mentioned in section~\ref{sec:mtp_warmup}, we adopt a cold-start setting for both our proposed method and baselines, only warming up the MTP module before the training process of MPO. 
See Appendix~\ref{app:implementation} for more implementation details.

\subsection{Task Performance}


We conduct experiments to evaluate the performance of MPO. The results are shown in Figure~\ref{fig:main_result} and Table~\ref{tab:humaneval}. The proposed method consistently outperforms the baseline approaches—including PPO, GRPO, and DAPO—across both GSM8K and MATH benchmarks, under both evaluated model architectures and scales.
This improvement demonstrates that optimizing over block-level semantic actions is beneficial for the model.
On the HumanEval benchmark (as shown in Table~\ref{tab:humaneval}), MPO also achieves steady gains over PPO, GRPO, and DAPO, confirming that its advantages are not limited to symbolic reasoning but also extend to program synthesis tasks.
Therefore, in code generation tasks, considering multi-step semantic dependencies plays a crucial role in ensuring both accuracy and coherence in generation.
Although the performance of MPO on GSM8K with a 7B model is close to the baseline, it achieves better results on more challenging benchmarks, such as MATH and HumanEval, indicating its advantage in structured reasoning tasks with larger models.

\subsection{Analysis of Training Stability}

\begin{figure*}[t]
    \captionsetup{aboveskip=3pt, belowskip=0pt}
    \centering
    \begin{subfigure}[t]{0.48\textwidth}
        \captionsetup{aboveskip=2pt, belowskip=0pt}
        \centering
        \includegraphics[width=\linewidth]{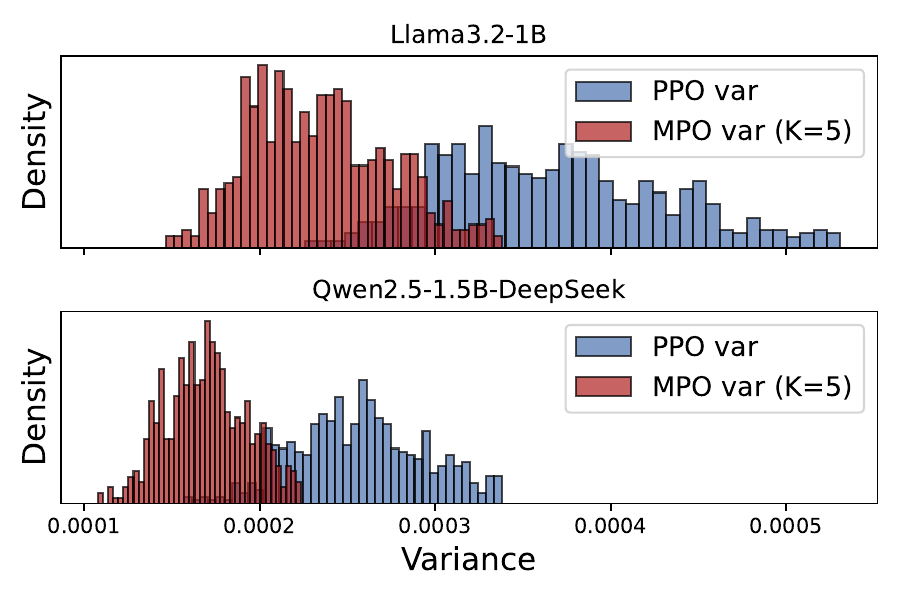}
        \caption{Variance of importance sampling ratio.}
        \label{fig:variance}
    \end{subfigure}
    \hfill
    \begin{subfigure}[t]{0.48\textwidth}
        \captionsetup{aboveskip=2pt, belowskip=0pt}
        \centering
        \includegraphics[width=\linewidth]{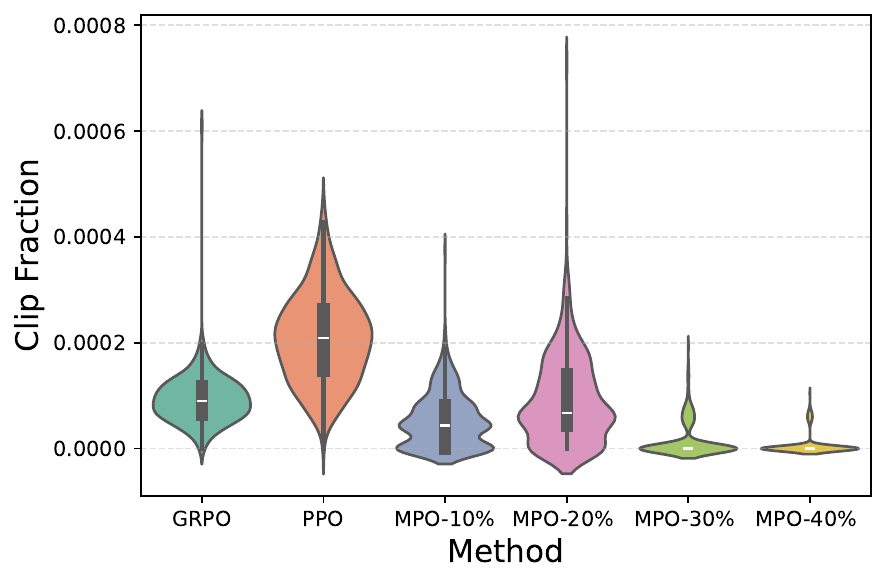}
        \caption{IS Ratio Clip fraction among strategies.}
        \label{fig:IS}
    \end{subfigure}
    \caption{Comparison of the variance of importance sampling ratios and clip fraction during training.} 
    \label{fig:two_plots}
    \vspace{-5pt}
\end{figure*}

\begin{figure*}[t]
    \captionsetup{aboveskip=3pt, belowskip=0pt}
    \centering
    \captionsetup{type=table}
    \begin{subfigure}[t]{0.47\textwidth}
        \centering
        \begin{tabular}{l|lll}
        \hline
        \multirow{2}{*}{Dataset} & \multicolumn{3}{c}{MPO Task Accuracy} \\ \cline{2-4} 
                                 & K=2         & K=3        & K=5        \\ \hline
        GSM8K                    & 0.871       & 0.875      & \textbf{0.882}      \\ \hline
        MATH                     & 0.771       & 0.753      & \textbf{0.789}      \\ \hline
        \end{tabular}
        \caption{Analysis of the number of MTP modules.}
        \label{tab:num_K}
    \end{subfigure}
    \hfill
    \begin{subfigure}[t]{0.49\textwidth}
        \begin{tabular}{c|ccc}
        \hline
        MTP Weight $\beta_2=$   &$0.08$ & $0.06$   & $0.04$   \\ \hline
        $\lambda=1.0$                        & 0.745   & 0.773   & \textbf{0.787}  \\ \hline
        $\lambda=0.9$ & 0.756   & 0.785   & 0.760  \\ \hline
        $\lambda=0.8$  & \textbf{0.758}   & \textbf{0.789}  & 0.764  \\ \hline
        \end{tabular}
        \caption{Grid Search of beta and decay rate.}
        \label{tab:grid}
    \end{subfigure}
    \caption{Effect of MTP block size $K$ and decay rate $\lambda$ on training stability and performance. Extending the block size to $K=5$ and applying a moderate decay \(\lambda=0.8\) produces the most stable and effective result.}
    \label{fig:two_tables}
    \vspace{-10pt}
\end{figure*}


As discussed in Section~\ref{sec:mul-obj}, incorporating block-wise information broadens the optimization horizon of each action and helps reduce bias, particularly in mathematical reasoning and code generation tasks where semantic units naturally span multiple tokens.
Empirically, we observe that MPO stabilizes the importance sampling ratio throughout training. As shown in Figure~\ref{fig:variance} and Figure~\ref{fig:IS}, analysis on GSM8K reveals that integrating multi-token information consistently lowers both the variance of the importance sampling ratio and the clip fraction during optimization. Notably, this stabilizing effect becomes more pronounced as the weights $\beta_k$ associated with the MTP modules increase.
\subsection{Ablation Study of MPO Weights}
\label{sec:proportion}
\begin{figure}[htbp]
  \centering
  \includegraphics[width=\linewidth]{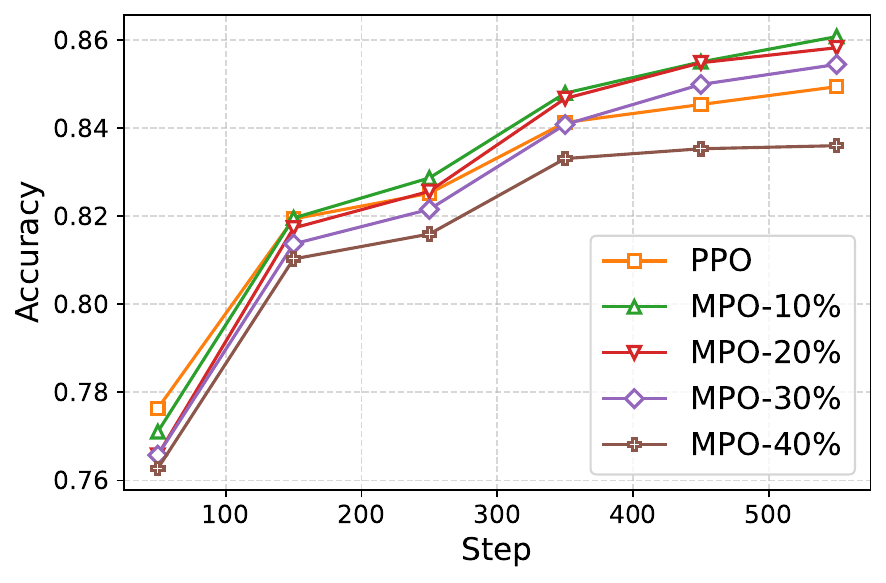}
  \caption{The effect of varying the proportion of weights incorporated from the MTP modules.}
  \label{fig:mtp_weights}
  \vspace{-15pt}
\end{figure}

Another important question is how increasing the contribution of MTP weights influences learning dynamics. As shown in Figure~\ref{fig:mtp_weights}, we examine the effect of varying the proportion of multi-token information incorporated via MTP modules on the GSM8K dataset using the DeepSeek-Distill-Qwen‑1.5B model.
The block-level contribution is controlled by adjusting the cumulative weights of the MTP modules (i.e., the sum of $\beta_2$ to $\beta_K$), while holding the decay factor fixed at $\lambda=0.8$ as defined in Eq.~\ref{eq:decay}.
Increasing the proportion of future information initially enhances policy stability, as evidenced by reduced variance in importance sampling ratios and a lower frequency of gradient clipping.
Optimal performance is achieved when approximately 10\%~20\% of the training signal originates from MTP modules.
However, when this proportion becomes too large, the adopted ratio approximation and the noise introduced by multi-token prediction increase bias, ultimately weakening the alignment between gradient updates and the next-token action objective.

\begin{figure*}[t]
    \captionsetup{aboveskip=3pt, belowskip=0pt}
    \centering
    \begin{subfigure}[t]{0.48\textwidth}
        \captionsetup{aboveskip=2pt, belowskip=0pt}
        \centering
        \includegraphics[width=\linewidth]{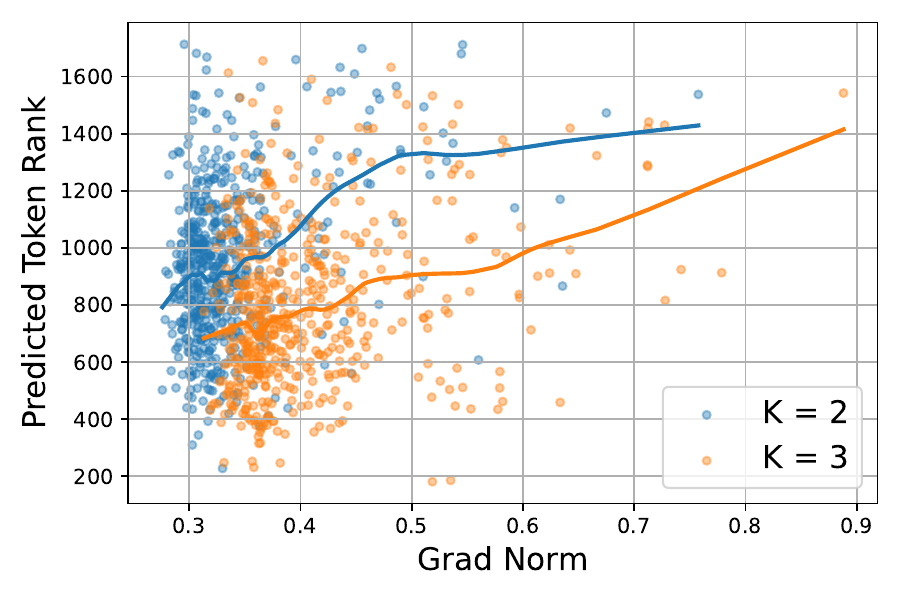}
        \caption{Predicted token rank with avg. gradient norm.}
        \label{fig:rank_grad}
    \end{subfigure}
    \hfill
    \begin{subfigure}[t]{0.48\textwidth}
        \captionsetup{aboveskip=2pt, belowskip=0pt}
        \centering
        \includegraphics[width=\linewidth]{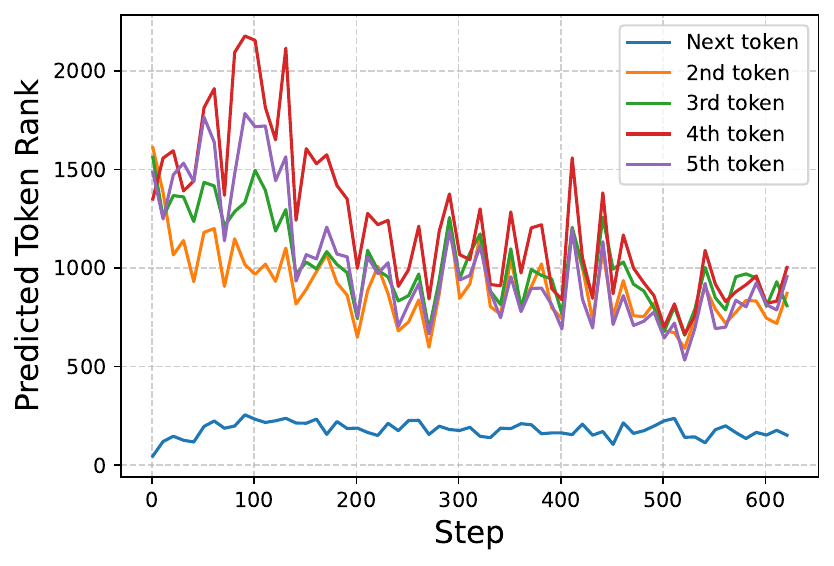}
        \caption{Predicted token rank with training steps.}
        \label{fig:rank_step}
    \end{subfigure}
    \caption{Reliability analysis of the multi-token information.
(a) Larger prediction errors correlate with higher gradient magnitudes.
(b) Progressive improvement in the MTP module’s next-K-token prediction capability.} 
    \label{fig:rank_plots}
    \vspace{-10pt}
\end{figure*}

\subsection{The Bias–variance Trade‑off in MPO} 
Having analyzed the stability of MPO under moderate integration of multi-token information, we conducted a grid search over the number of aggregated tokens $K$ and the decay rate $\lambda$ to assess their effects on training stability and optimization performance.
As shown in Table~\ref{tab:num_K}, increasing $K$ up to 5 leads to improved performance, suggesting that extending the decision horizon within this range is beneficial; however, larger $K$ values tend to introduce additional noise.
Table~\ref{tab:grid} summarizes the effects of varying the decay rate $\lambda$ and the cumulative MTP weight $\sum_{k=2}^{K}\beta_k$. Across all experiments, we keep the initial weight $\beta_2$ small so that the total block-level contribution remains below 30\%, which stabilizes training while preserving meaningful multi-token context. If $\beta_2$ is too small (e.g., $\beta_2 = 0.04$), the stability of information from the MTP modules declines. In general, applying a decay term for more distant predictions improves stability, as appropriately discounting further predictions helps constrain variance while maintaining a sufficiently extended decision horizon—balancing bias reduction and training stability in MPO.

\subsection{Noise Analysis of Multi-token Prediction}
As discussed in section~\ref{sec:proportion}, excessive MTP weights were found to reduce overall performance. We further examine whether such instability arises from noisy multi-token information. We use the logit rank of the predicted token—its position within the backbone model’s vocabulary distribution—as a proxy for prediction accuracy.
Figure~\ref{fig:rank_grad} shows the relationship between the $2^{nd}$‑token prediction rank and the average batch gradient norm under $K=2$ and $3$. 
Higher ranks correlate with larger gradient norms, suggesting that less accurate multi-token predictions can indeed amplify gradient noise and destabilize training.
Figure~\ref{fig:rank_step} further illustrates that, although only warmed up initially, the MTP module’s predictions improve steadily during training even without explicit entropy regularization. 
The impact of prediction noise remains controlled.


\subsection{Efficiency Analysis}


\noindent\textbf{Training Efficiency.} Table~\ref{table:speed} and Table~\ref{table:efficiency} summarize the resource overhead and efficiency of MPO on a single node with \(8\times\) NVIDIA A100 (80 GB). For Table~\ref{table:efficiency} we use Deepseek-Distill-Qwen2.5-1.5b as the base model, and we set group size=4 for GRPO and K=5 for MPO. The memory consumption of MPO is comparable to that of GRPO, while training runs approximately 30\% faster on average. MPO contributes controlled training burden while offering substantial performance benefits. The additional parameters introduced by MTP are used solely during training to estimate multi-token policy ratios and are detached during inference, ensuring a fair comparison with baseline methods. 

\noindent\textbf{Warm-up Strategy.} In MPO, we include a brief warm‑up phase; this design choice follows prior multi-head decoding frameworks\citep{cai2024medusa,ankner2024hydra}. The warm‑up phase costs approximately 12 minutes while PPO or MPO training costs approximately 7 hours for 20 epochs on average. The additional time overhead is negligible compared to the main training stage. More discussion about efficiency is in Appendix~\ref{app:efficiency}.

\begin{table}[]
\centering
\begin{tabular}{lccc}
\hline
\multicolumn{4}{l}{Training Speed (Second per iteration)}                                   \\ \hline
\multicolumn{1}{l|}{Model}                         & PPO  & K=3  & K=5  \\ \hline
\multicolumn{1}{l|}{Deepseek-Qwen2.5-1.5b} & 25.6 & 30.3 & 33.3 \\
\multicolumn{1}{l|}{Llama3.2-1b}          & 19.1     & 25.0     & 28.3     \\ \hline
\end{tabular}
\caption{The average training time with block size K.}
\label{table:speed}
\vspace{-5pt}
\end{table}

\begin{table}[t] 
\centering
\begin{tabular}{l|c|c}
\hline
Method & Memory Usage & Time Per Step \\
\hline
PPO & 22.1\% & 25.6 s\\
GRPO & 30.6\% & 43.7 s \\
MPO & 31.0\% & 33.3 s \\
\hline
\end{tabular}
\caption{Memory and time efficiency.}
\label{table:efficiency}
\vspace{-15pt}
\end{table}

\section{Conclusion}

In this paper, we propose Multi‑Token Policy Gradient Optimization (MPO), mitigating the gap between token-level optimization and the structured reasoning behavior required by complex reasoning tasks.
Experiments on GSM8K, MATH, and HumanEval show that MPO consistently outperforms standard policy‑gradient baselines. Further analyses demonstrate that moderate incorporation of multi-token information stabilizes training by reducing gradient variance, and that there exists an optimal integration ratio balancing bias variance trade-off.
Our paper not only introduces a practical multi-token policy gradient algorithm, but also marks a methodological shift that moves beyond token-level optimization in the post-training of LLMs.
We hope this study inspires future research to further explore the granularity of policy learning and advance toward semantically structured decision-making process in the LLMs.

\section*{Limitations}
\label{app:future}

MPO builds upon Multi-Token Prediction (MTP), a technique primarily explored in pre-training and speculative decoding. Its application to post-training optimization remains understudied, presenting several open research questions:

\noindent\textbf{1. Warmup Method.} While MTP-based training has been shown to improve reasoning (DeepSeek V3 and subsequent work), applying it during post-training—especially to models not pre-trained with MTP (e.g., Llama-3, Qwen2.5)—may degrade final performance after reinforcement learning. Although MPO mitigates this by freezing the backbone, developing more effective warmup strategies remains an open challenge.

\noindent\textbf{2. Scaling Block Size.} MPO incurs computational overhead as the number of MTP modules increases. Our experiments used up to K=5 modules, whereas prior work suggests larger K (e.g., 8) may yield further gains. Designing efficient methods to leverage longer future contexts is crucial, especially for scaling block-level policy optimization.

\noindent\textbf{3. Alternative Optimization Objectives.} This work implements MPO based on a PPO-based objective. As noted in Section~\ref{sec:mul-obj}, MPO could also be integrated with other objectives (e.g., GRPO and it variants), but implementing MPO based on GRPOs may encounter challenges like rebalancing the bias-variance trade-off. Investigating how MPO interacts with different surrogate objectives presents a promising direction for future work.

\bibliography{custom}

\appendix

\section{Implementation Details}
\label{app:implementation}

We will provide separate explanations of the overall MPO implementation process, the experimental parameter settings, and details of the code and hardware implementation.

\subsection{MTP Modules}

\begin{figure*}[h]
    \centering    
    \includegraphics[width=1\linewidth]{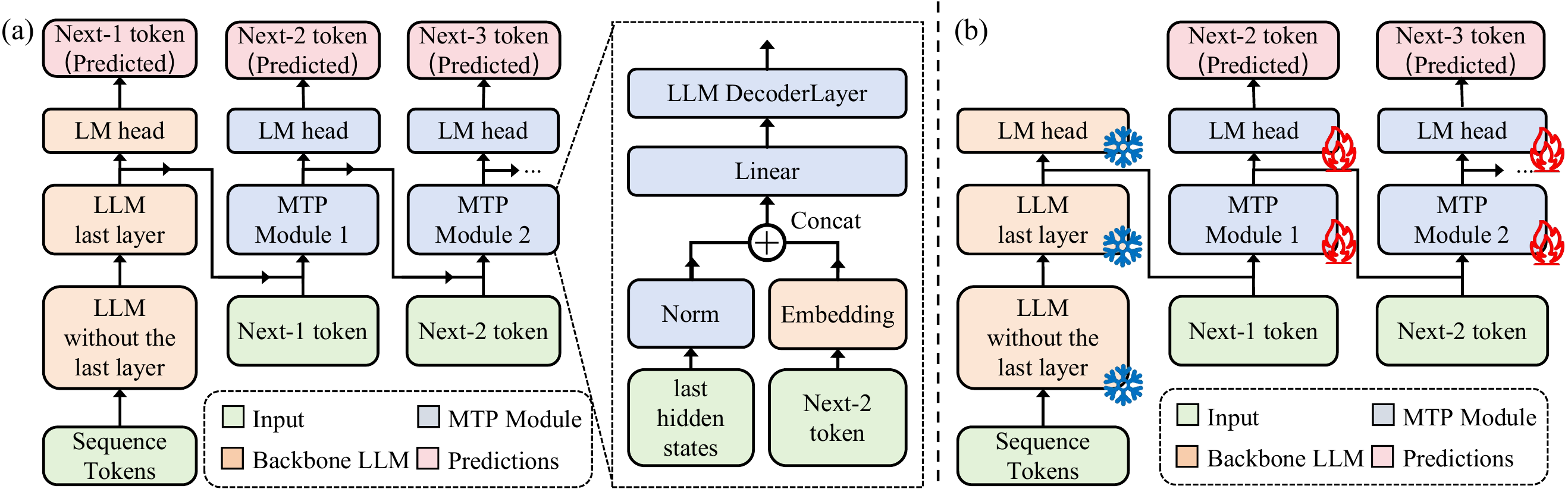}
    \caption{(a) The implementation of the MTP module used in MPO, the structure preserves causal consistency following Deepseek-V3; (b) Demonstration of the MPO warmup stage to initialize MTP Modules before the RL stage.}
    \label{fig:warmup}
\end{figure*}

\begin{table*}[htbp]
    \centering
    \begin{tabular}{lccc}
    \toprule
    Dataset & \# Train Samples & \# Test Samples & Avg. Sample Length \\
    \midrule
    GSM8K   & 7,473     & 1,319    & $\sim$80 tokens \\
    MATH    & 7,500     & 5,000    & $\sim$200 tokens \\
    HumanEval    & 1399 (MBPP)     & 164    & $\sim$130 tokens \\
    \bottomrule
    \end{tabular}
    \caption{Statistics of datasets used for training and evaluation.}
    \label{tab:data_stats}
\end{table*}

The implementation of MPO requires the MTP module to predict the model’s policy distributions at multiple positions along a given trajectory $i$, enabling the calculation of importance sampling coefficients over blockwise sequences. Achieving stable multi-position decision making depends on the MTP module’s ability to produce accurate predictions of multi-token policy distributions. These distributions cannot be directly modeled using only the backbone’s final layer and language output head, so an appropriate warmup is necessary. Although previous works has noted that MTP can improve the model’s reasoning ability during fine-tuning~\citep{liu2024deepseek}, to ensure fair comparisons, we do not backpropagate gradients to the backbone during MPO finetuning and only train the MTP module (as shown in Figure~\ref{fig:warmup}), which has also been adopted in previous efficient MTP decoding works~\citep{cai2024medusa}. As we discussed previously, each MTP module and its corresponding LM head are deeply copied from the backbone model’s final layer and LM head, respectively, to avoid an overly complex initialization process.

\subsection{Dataset and Training Parameters}
\label{app:data_details}

We conduct experiments using three widely used mathematical reasoning and coding benchmarks: GSM8K~\citep{cobbe2021training}, MATH~\citep{hendrycks2021measuring}, and HumanEval~\citep{chen2021evaluating}. For HumanEval, we combined the sanitized and original data provided in coding benchmark MBPP~\citep{austin2021program} to conduct MPO training.
Table~\ref{tab:data_stats} summarizes the main statistics of these datasets. The question-answer pairs in both datasets are provided in a natural language format, suitable for language model-based reasoning experiments. The open source mathematical and coding benchmarks used do not contain any private information or offensive content.

We shuffle and concatenate the training splits of two math datasets for MTP warm-up training. For MPO training, we conduct separate experiments on each dataset, using the corresponding training split for learning and the corresponding test split for evaluation. This setup allows us to evaluate the generalization and effectiveness of our method on both mathematical reasoning and coding tasks.

\subsection{MPO Training Process} 

\begin{table*}[htbp]
    \centering
    \begin{tabular}{lc}
    \toprule
    Parameter                   & Value \\
    \midrule
    Train batch size            & 256 \\
    PPO mini batch size         & 128 \\
    PPO micro batch size / GPU  & 4    \\
    Number of GPUs               & 8    \\
    Number of Nodes             & 1    \\
    Max prompt length           & 1024  \\
    Max response length         & 2048  \\
    Learning rate (actor)       & $1\times10^{-6}$ \\
    Learning rate (critic)      & $1\times10^{-5}$ \\
    Total epochs                & 20 \\
    \\[-1em]
    \bottomrule
    \end{tabular}
    \caption{Key hyperparameters for MPO training.}
    \label{tab:mpo_train_params}
\end{table*}

\begin{algorithm*}[htbp]
\caption{Multi-Token Policy Optimization (MPO)}
\label{alg:mtp}
\SetKwInOut{Input}{Input}
\SetKwInOut{Output}{Output}

\Input{
    Dataset $\mathcal{D}$ of prompts $q_i$ and annotated outputs $o'_i$.
    Pretrained LLM with MTP module; initial parameters $\theta$.
    Old policy parameters $\theta_\mathrm{old}$.
    Block size $K$, weights $\alpha_k$, $\beta_k$; clip parameter $\epsilon_{low},\epsilon_{high}$
}
\Output{Updated policy parameters $\theta$}

\BlankLine
\textbf{Stage 1: MTP Warm-Up} \\
\For{several epochs}{
    \For{batch of $(q_i, o'_i)$ in $\mathcal{D}$}{
        \For{each position $t$ in $o'_i$}{
            \For{$k = 1$ to $K$}{
                Predict $p(o'_{t+k} \mid q,o'_{1: t})$ using backbone model ($k=1$) and MTP module ($k\ge2$)
            }
            Compute $\mathcal{L}_\text{MTP} = -\sum_{k=2}^{K} \alpha_k \log p(o'_{t+k} \mid q,o'_{1: t})$
        }
        Update only MTP parameters by minimizing $\mathcal{L}_\text{MTP}$
    }
}

\BlankLine
\textbf{Stage 2: MPO Fine-tuning} \\
\For{each MPO epoch}{
    $\mathcal{B} \leftarrow$ sample mini-batch of prompts $q_i$ from $\mathcal{D}$ \\
    \For{each $q_i$ in $\mathcal{B}$}{
        Generate trajectory $o_{i}$ using $\pi_{\theta_\mathrm{old}}$ \\
        Compute per-token or group-based advantages $\hat{A}_{i,t}$
    }

    \For{each trajectory $(q_i, o_i)$ and each position $t$ in $o_i$}{
        \tcp{Sample $K$-step target block}
        Sample $o_{i, t+1:t+K}$ in an autoregressive manner using $\pi_{\theta_\mathrm{old}}$ \\
        Initialize $D_{kl} = 0$ \\
        \For{$n = 1$ to $K$}{
            Compute: \\
            $\;\; p_\theta = \pi_\theta(o_{i,t+n} | o_{i,1:t+n-1})$ \\
            $\;\; p_{\theta_\mathrm{old}} = \pi_{\theta_\mathrm{old}}(o_{i,t+n} | o_{i,1:t+n-1})$ \\
            $\;\; D_{kl} += \beta_n \cdot (\log p_\theta - \log p_{\theta_\mathrm{old}})$
        }
        $\widetilde{R}_{i,t}^{(K)} = \exp(D_{kl})$ \\
        Compute surrogate loss for $(i, t)$: \\
        $J_{i,t} = \min \left( \widetilde{R}_{i,t}^{(K)} \hat{A}_{i,t}, \;\; \text{clip}(\widetilde{R}_{i,t}^{(K)}, 1-\epsilon_{low}, 1+\epsilon_{high}) \hat{A}_{i,t} \right)$
    }

    Aggregate loss:
    $J_\mathrm{MPO} = \frac{1}{N} \sum_{i} \sum_{t} J_{i,t}$.

    Update parameters $\theta$ via gradient descent using $J_\mathrm{MPO}$.
    Optionally, update $\theta_\mathrm{old}$ periodically.
}
\end{algorithm*}

Here we provide additional implementation details for the Multi-Token Policy Optimization (MPO) method described above, including the hyperparameters and a pseudocode Algorithm~\ref{alg:mtp} for clarity. Table~\ref{tab:mpo_train_params} lists our main hyperparameters and settings for the MPO training process. These values are extracted from our training scripts and reflect typical settings for large-scale RLHF or policy optimization tasks on math reasoning datasets. For HumanEval, due to the extensive thinking behavior of Deepseek distilled models, we extend the response length to 3072.

\subsection{Potential Risks} 
Our proposed MPO method is based on reinforcement learning. As discussed in the main text, high multi-token weights may cause training instability and model divergence; thus, care must be taken to maintain training stability. Since our experiments are limited to mathematical and coding reasoning tasks, there are no additional risks of privacy leakage or offensive content. Regarding data and models, we only use open- source datasets consisting of mathematical and programming problems, containing no personal or sensitive data. The employed models are commonly used open-source LLMs, whose potential behavioral risks are unrelated to our research topic.

\section{Discussion of Multi-Token Approach}
\label{app:mtp_formal}


This appendix provides a formal discussion of why incorporating
Multi-token prediction into policy optimization can improve optimization
performance and enhance long-horizon reasoning ability. The exposition is
consistent with the notation and definitions in the main text
(Sections~\ref{sec:mtp_warmup}--\ref{sec:future}).

\subsection{Intuitive Example and Motivation}

In standard policy optimization methods such as PPO, the policy improvement step is based on per-token decisions, where the importance ratio
\begin{equation}
r_{i,t}(\theta) = \frac{\pi_\theta(o_{i,t+1}\mid q_i,o_{i,1:t})}{\pi_{\theta_{\text{old}}}(o_{i,t+1}\mid q_i,o_{i,1:t})}
\end{equation}
compares the likelihood of a single action (next token) under the updated and old policies.  
This formulation implicitly assumes that:  
1) the reward structure can be well captured by local token-level updates, and  
2) long-horizon dependencies are either negligible or already encoded in the advantage estimator $\hat A_{i,t}$.

While traditional one-step token-level policy optimization updates the model based on individual token predictions, our illustration highlights that semantics and reasoning patterns in language generation typically emerge from coherent blocks of multiple tokens. As shown in the figure, generating a valid reasoning step or factual entity (e.g., \$a=5\$ or \$b = a\^{}2 = 25\$) requires accurately predicting a sequence of tokens as a whole.

Limiting optimization to one‑step advantage estimation introduces several issues:
\begin{enumerate}
\item \textbf{Fragmented learning signals}: Token‑level updates isolate local token actions, weakening the internal consistency of reasoning segments.
\item \textbf{Incomplete credit assignment}: Single‑token objectives struggle to capture dependencies whose effects span across multiple tokens within a reasoning step.
\item \textbf{Granularity mismatch}: The optimization signal operates at the token level, while the semantic correctness of reasoning typically emerges at the segment level.
\end{enumerate}

In contrast, as illustrated in the lower MTP/MPO section, aggregating actions and optimization targets at the block level enables updates that better capture semantic integrity, reduce estimation variance, and align training with the goals of structured reasoning.

By instead defining a $K$-step decision ratio,
\begin{equation}
R_{i,t}^{(K)}(\theta) = \prod_{n=1}^{K}\frac{\pi_\theta(o_{i,t+n}\mid o_{i,1:t+n-1})}{\pi_{\theta_{\text{old}}}(o_{i,t+n}\mid o_{i,1:t+n-1})},
\label{eq:ratio}
\end{equation}
we directly compare the joint likelihood of a future span under the new and old policies.  
This transforms the optimization from purely myopic next-token correction into a multi-step decision process, allowing the policy gradient to align with behaviors that unfold across multiple tokens.  
In essence, the surrogate objective is no longer tied to local moves but reflects higher-level planning, where corrections propagate over spans of length $K$ rather than being confined to single steps.

\subsection{Theoretical Derivation and Bias Reduction}
\label{app:math_prove}

\subsubsection{Problem Setup and Notation}

We consider a standard reinforcement learning setup for sequence generation. Let time steps be indexed by \(t = 0,1,2,\dots\), where each step corresponds to generating a token.  

\begin{itemize}
    \item State: \(s_t\) denotes the state at time \(t\), which includes the context (previous tokens and hidden states).
    \item Action \(a_t\) denotes the token generated at time \(t\).
    \item Policy: \(\pi_\theta(a_t \mid s_t)\), with \(\pi_{\text{old}}\) denoting the previous policy.
    \item Discount factor: \(\gamma \in (0,1]\).
    \item Value function: \\\(V^\pi(s) = \mathbb{E}_\pi \left[ \sum_{k=0}^\infty \gamma^k r_{t+k} \,\big|\, s_t = s \right]\)
    \item Advantage function: \\\(A^\pi(s_t, a_t) = Q^\pi(s_t,a_t) - V^\pi(s_t)\)
    \item Estimated value function: \(\hat V_\phi(s)\)
\end{itemize}

In standard one-step PPO, the advantage is often estimated using the TD error and generalized advantage estimator (GAE):
\begin{align}
    \delta_t &= r_t + \gamma \hat V_\phi(s_{t+1}) - \hat V_\phi(s_t), \\
    \hat A^{\text{GAE}}_t &= \sum_{l=0}^{\infty} (\gamma \lambda)^l \delta_{t+l},
\end{align}
where \(\lambda \in [0,1]\) controls the trade-off.

\subsubsection{K-Step / Multi-Token Core Idea}

In K-step (multi-token) prediction, at time \(t\) we consider predicting not just \(a_t\), but the next \(K\) tokens jointly. Define the K-step return:
\begin{equation}
    G^{(K)}_t := \sum_{k=0}^{K-1} \gamma^k r_{t+k} + \gamma^K \hat V_\phi(s_{t+K}).
\end{equation}
The corresponding K-step advantage estimator is
\begin{equation}
    \hat A^{(K)}_t := G^{(K)}_t - \hat V_\phi(s_t).
\end{equation}
Intuitively, as \(K\) increases, the advantage relies more on the actual observed rewards and less on the bootstrapped value function \(\hat V_\phi\), thereby reducing the bias from value function approximation.

\subsubsection{Bias Analysis}

Let \(\epsilon_V := \sup_s \left| \hat V_\phi(s) - V^\pi(s) \right|\) denote the maximum value function approximation error. The K-step return based on the true value function is
\begin{equation}
    G^{(K),\text{true}}_t := \sum_{k=0}^{K-1} \gamma^k r_{t+k} + \gamma^K V^\pi(s_{t+K}),
\end{equation}
so the error due to \(\hat V_\phi\) is
\begin{equation}
\begin{split}
    \Delta G^{(K)}_t :&= G^{(K)}_t - G^{(K),\text{true}}_t \\&= \gamma^K (\hat V_\phi(s_{t+K}) - V^\pi(s_{t+K})).
\end{split}
\end{equation}
Hence, in absolute value,
\begin{equation}
    |\Delta G^{(K)}_t| \le \gamma^K \epsilon_V.
\end{equation}
The bias of the K-step advantage relative to the true K-step advantage is
\begin{equation}
\begin{split}
        \hat A^{(K)}_t - A^{\text{true},(K)}_t &= 
    \big(G^{(K)}_t - \hat V_\phi(s_t)\big) \\&\;\;\;\;\;\;- \big(G^{(K),\text{true}}_t - V^\pi(s_t)\big) \\
    &= \gamma^K (\hat V_\phi(s_{t+K}) - V^\pi(s_{t+K})) \\&\;\;\;\;\;\;- (\hat V_\phi(s_t)-V^\pi(s_t)),
\end{split}
\end{equation}
with absolute value bounded by
\begin{equation}
    \left| \hat A^{(K)}_t - A^{\text{true},(K)}_t \right| \le \gamma^K \epsilon_V + \epsilon_V = (1+\gamma^K) \epsilon_V.
\end{equation}

If we focus on the relative dependence on one-step (n=1), note that the error term for one-step is $(1+\gamma)\epsilon_V$. More importantly, if we decompose the bias into ``the bootstrapping bias from the tail'' and ``the bias from the current estimate,'' we can see that the bootstrapping term from the tail is multiplied by $\gamma^K$.

In particular, when we consider the bias propagation with respect to the true infinite-horizon return, the bias term related to n-step bootstrapping decays as $\mathcal{O}(\gamma^K)$. In other words, the bias is introduced by bootstrapping (relying on $\hat{V}_\phi(s_{t+n})$), and its coefficient is $\gamma^K$. As $n$ increases, this term decays exponentially by a factor of $\gamma^K$, thus reducing reliance on $\hat{V}_\phi$ $\Rightarrow$ reducing the bias introduced by function approximation.

\subsubsection{Expected Advantage Bias}

Consider the expected bias over trajectories:
\begin{equation}
\begin{aligned}[t]  
&\mathbb{E}\big[ \hat A^{(K)}_t - A^\pi(s_t, a_t) \big] 
\\&= \mathbb{E}\Big[ G^{(K)}_t - \hat V_\phi(s_t) - (Q^\pi(s_t,a_t) - V^\pi(s_t)) \Big] \\
&= \mathbb{E}\Big[ \sum_{k=0}^{K-1} \gamma^k r_{t+k} + \gamma^K \hat V_\phi(s_{t+K}) - \hat V_\phi(s_t) \\&\quad- \sum_{k=0}^{\infty} \gamma^k r_{t+k} + V^\pi(s_t) \Big] \\
&= \mathbb{E}\Big[ \gamma^K (\hat V_\phi(s_{t+K}) - V^\pi(s_{t+K})) \\&\quad- (\hat V_\phi(s_t)-V^\pi(s_t)) \\&\quad - \sum_{k=K}^{\infty} \gamma^k (r_{t+k} - \mathbb{E}[r_{t+k} \mid s_{t+K}]) \Big].
\end{aligned}
\end{equation}

\begin{table*}[h]
\centering
\renewcommand{\arraystretch}{1.2}
\scalebox{0.87}{
\begin{tabular}{l|lll|lll}
\hline
\multicolumn{1}{c|}{\multirow{2}{*}{Settings}} & \multicolumn{3}{c|}{GSM8K}                                                   & \multicolumn{3}{c}{MATH}                                                    \\ \cline{2-7} 
\multicolumn{1}{c|}{}                          & \multicolumn{1}{c}{K=2} & \multicolumn{1}{c}{K=3} & \multicolumn{1}{c|}{K=5} & \multicolumn{1}{c}{K=2} & \multicolumn{1}{c}{K=3} & \multicolumn{1}{c}{K=5} \\ \hline
MPO                              &0.871                         &0.875                         &\textbf{0.882}                        & 0.771                        & 0.753                        & \textbf{0.789}                        \\ \hline
+ $V_{MPO}$                                            &0.879                         &0.875                         & \textbf{0.882}                         &0.773                         & 0.756                        & 0.762                        \\ \hline
\end{tabular}
}
\caption{MPO performance with different token block size K and ablation study of multi-token value estimation.}
\label{tab:ablation}
\vspace{-10pt}
\end{table*}

Under standard assumptions, the last term has zero expectation, so the dominant contribution to bias comes from
\begin{equation}
\begin{split}
        &\mathbb{E}\big[ \hat A^{(K)}_t - A^\pi(s_t, a_t) \big] \\&\approx 
    \mathbb{E} \Big[ \gamma^K (\hat V_\phi(s_{t+K}) - V^\pi(s_{t+K})) \\&\quad- (\hat V_\phi(s_t)-V^\pi(s_t)) \Big].
\end{split}
\end{equation}

If the bias of $\hat{V}_\phi$ at different time steps is not systematically correlated (or its expectation is approximately zero), then the main remaining error comes from the tail error scaled by $\gamma^n$, with the overall expected bias reduced to $\mathcal{O}(\gamma^n \epsilon_V)$. The expected bias due to reliance on the bootstrapped value in n-step is reduced to the order of $\gamma^n$, thereby lowering the bias.

\section{Multi-token Value Estimation}
\label{app:value}

\subsection{Multi-Token Value Heads}

\begin{figure}[h]
    \centering
    \includegraphics[width=0.8\linewidth]{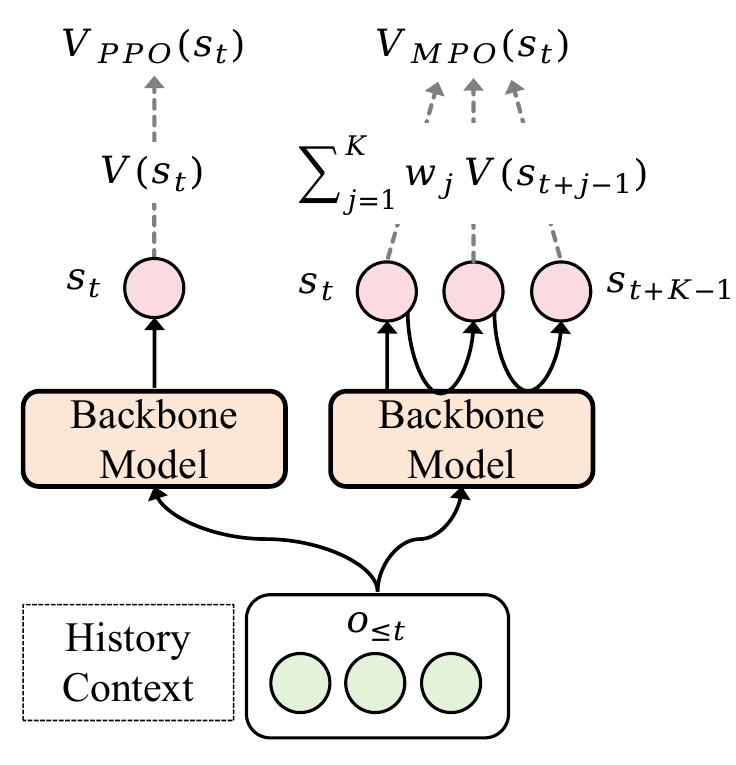}
    \vspace{-8pt}
    \caption{The implementation of the value function on the original auto-regressive backbone model (left) and those with the MTP module (right). For multi-token prediction models, we try a weighted sum of the output from the shared value head.}
    \label{fig:value_head}
\end{figure}

In parallel with our multi-token perspective modification for the actor, we have also conducted experiments on the critic model, allowing it to anticipate the returns of actions over a future interval during training. To accommodate multiple multi-token prediction heads, we aggregate their outputs at each decision point via a learned convex combination:
\begin{equation}
\begin{split}
V_{\text{MPO}}(s_t) = \sum_{j=1}^{K} w_{j} V_{j}(\hat s_{t+j-1})
\end{split}
\end{equation}
where $\hat s_{t+j-1}$ denotes the predicted states and
\begin{equation}
\begin{split}
w_{j} = \frac{\exp(\gamma_j)}{\sum_{k=1}^K \exp(\gamma_k)}, \quad \boldsymbol{\gamma} = (\gamma_1, \ldots, \gamma_K)
\end{split}
\end{equation}
with $\gamma$ as learnable parameters, balancing the value obtained from predicted states. $V_{j}$ is the value head append to the $j^{th}$ MTP module. Given this value estimation, the advantage at each position $t$ is computed as
\begin{equation}
\begin{split}
\hat{A}_{i,t} = \hat{Q}_{i,t} - V_{\text{MPO}}(s_t),
\end{split}
\end{equation}
where $\hat{Q}_{i,t}$ is the estimated return as in standard policy gradient methods. This approach permits each value head to specialize, while the aggregation provides a flexible baseline for stable advantage calculation and policy optimization over multi-token predictions. Figure~\ref{fig:value_head} illustrates the value estimation process.

\subsection{Effect of Block size on Value Estimation}
\label{sec:mtp_num}

‌We conducted experiments with Deepseek-Distill-Qwen2.5-1.5b by fixing the weight sum of MTP modules ($\sum_{k=2}^{K}\beta_k$) of knowledge at 20\%, with results shown in Table~\ref{tab:ablation}. The method of injecting multi-token information from the critic side did not improve overall performance as expected. Although our initial experiments suggested that this approach could have a "symmetric" effect—namely, by training the MTP module on the value prediction task to enhance overall training stability—this was not supported by later results. After addressing certain implementation issues in our code, it appears that MPO is not sensitive to the incorporation of multi-token information into the value function. In fact, with K = 4 or 5, it may introduce greater instability and cause the model to oscillate at a suboptimal level. Simply injecting value predictions by summing the outputs of multiple MTP modules does not seem to effectively improve value estimation accuracy.

\section{Discussion of Efficiency}
\label{app:efficiency}

\begin{table}[]
\renewcommand{\arraystretch}{1.15}
\scalebox{0.9}{
\begin{tabular}{lccc}
\hline
\multicolumn{4}{l}{Training Speed (Second per iteration)}                                   \\ \hline
\multicolumn{1}{l|}{Model}                         & PPO  & K=3  & K=5  \\ \hline
\multicolumn{1}{l|}{Deepseek-Qwen2.5-1.5b} & 25.6 & 30.3 & 33.3 \\
\multicolumn{1}{l|}{Llama3.2-1b}          & 19.1     & 25.0     & 28.3     \\ \hline
\multicolumn{4}{l}{Model Size (Compared to the original model)}                                          \\ \hline
\multicolumn{1}{l|}{Model}                         & PPO  & K=3  & K=5  \\ \hline
\multicolumn{1}{l|}{Deepseek-Qwen2.5-1.5b} & 1.0$\times$ & 1.20$\times$ & 1.41$\times$ \\
\multicolumn{1}{l|}{Llama3.2-1b}          & 1.0$\times$     & 1.52$\times$     & 2.05$\times$     \\ \hline
\end{tabular}
}
\caption{Efficiency analysis result.}
\label{table:speed_1}
\vspace{-10pt}
\end{table}

\textbf{Computation overhead.} Table~\ref{table:speed_1} illustrates further analysis with the resource overhead and comparative efficiency of MPO. The current implementation of MTP in MPO does not k-fold increase the overall computational cost. Since all logits needed for multi-token ratios are obtained from a single standard forward pass, the MTP module generates predictions for K tokens in parallel within one forward pass, without requiring K separate model evaluations.
Despite this computational overhead for an additional forward pass used to compute multi-token policy ratios after obtaining full trajectories, MPO delivers consistent accuracy improvements on both GSM8K and MATH benchmarks over PPO and GRPO.
These results indicate that the added MTP modules contribute minimal training burden while offering substantial efficiency and performance benefits.

\noindent\textbf{Design of Warmup Stage.}
To ensure stable optimization, we include a brief warm‑up phase to pre‑train the MTP modules (approximately 12 minutes for one epoch) before the MPO fine‑tuning stage (around 7 hours for 20 epochs), incurring only a lightweight additional time cost.
This design choice follows prior multi-head decoding frameworks~\citep{cai2024medusa,ankner2024hydra} and represents a standard stabilization step rather than additional complexity unique to our method.

MPO emphasizes a modular and practical architecture; this design enables the framework to improve optimization stability and reasoning performance; furthermore, future work may explore even lighter-weight MTP variants for large‑scale deployment (also discussed in Limitations).

\section{Usage of Large Language Models}

During the writing of this paper, large language models were used solely as tools for:
\begin{itemize}
    \item \textbf{Grammar checking}: Asking the LLM to check whether there is any grammar mistakes in the given paragraph.
    \item \textbf{Translation}: Asking the LLM to provide a proper translation for certain phrases and expressions.
    \item \textbf{Readability improvement}: Asking the LLM to give suggestions of how to change the tone of our written paragraphs for fluency.
\end{itemize}

\section{Reproducibility}

We will briefly go through the tools and data we use in this paper for reproducibility. 

\noindent\textbf{MTP Modules.} We developed our Multi-token Prediction modules based on transformers (ver 4.51.1) and trl (ver 0.22.0), both frameworks are open-sourced by the HuggingFace community. We modify the modeling library file of the transformers framework to achieve multi-token prediction.

\noindent\textbf{MPO Approach.} The overall MPO approach is developed based on the open-source reinforcement learning framework Verl (volcano engine, ver 0.3.1). We modify the PPO Trainer Class and Actor workers Class library to compute and gather multi-token log probability distributions when computing the policy loss of MPO.

\noindent\textbf{Datasets and models.} We conduct our experiments with open-source datasets GSM8K and MATH for MTP warmup and MPO training. The models are available on the open-source AI community websites. 

\noindent\textbf{Code and implementations.} In the supplementary materials, we include a detailed README file that guides the reproduction process, and we provide our implementation of the causal MTP module as well as the core code computing MPO policy loss for reference.

\end{document}